\documentclass[sigconf]{acmart}
\usepackage{subcaption}
\usepackage{tikz}
\usepackage{microtype} 

\usepackage{pbox}

\def\BibTeX{{\rm B\kern-.05em{\sc i\kern-.025em b}\kern-.08emT\kern-.1667em\lower.7ex\hbox{E}\kern-.125emX}}
    
\setcopyright{none}
\acmConference[]{In submission to IFCS Cluster Benchmarking Challenge}
\acmYear{2019}
\acmBooktitle{submission}
\acmPrice{}
\acmDOI{} 
\acmISBN{}

\newcommand{\urlBibTeX}[1]{\url{#1}} 

\newif\ifdraft
\drafttrue
\ifdraft
	\newcommand{\xtai}[1]{\textcolor{blue}{[[Xiao Hui says: #1]]}}
	\newcommand{\kfrisoli}[1]{\textcolor{olive}{[[Kayla says: #1]]}}
\else
	\newcommand{\xtai}[1]{}
	\newcommand{\kfrisoli}[1]{}
\fi

\begin{document}
%
\title{Benchmarking Minimax Linkage}

%
\author{Xiao Hui Tai}
\affiliation{\institution{Carnegie Mellon University}}
\email{xtai@andrew.cmu.edu}
\author{Kayla Frisoli}
\affiliation{\institution{Carnegie Mellon University}}
\email{kfrisoli@andrew.cmu.edu}
\renewcommand{\shortauthors}{Tai et al.}

%
%
%
%
%
%
%

%
\renewcommand{\shortauthors}{Tai, Frisoli}

%
\begin{abstract}
        Minimax linkage was first introduced by Ao et al. \cite{ogMinimax} in 2004, as an alternative to standard linkage methods used in hierarchical clustering. Minimax linkage relies on distances to a prototype for each cluster; this prototype can be thought of as a representative object in the cluster, hence improving the interpretability of clustering results. Bien and Tibshirani analyzed properties of this method in 2011 \cite{Bien2011}, popularizing the method within the statistics community. Additionally, they performed comparisons of minimax linkage to standard linkage methods, making use of five data sets and two different evaluation metrics (distance to prototype and misclassification rate). In an effort to expand upon their work and evaluate minimax linkage more comprehensively, our benchmark study focuses on thorough method evaluation via multiple performance metrics on several
well-described data sets. We also make all code and data publicly available through an R package, for full reproducibility. Similarly to \cite{Bien2011}, we find that minimax linkage often produces the smallest maximum minimax radius of all linkage methods, meaning that minimax linkage produces clusters where objects in a cluster are tightly clustered around their prototype. This is true across a range of values for the total number of clusters ($k$). However, this is not always the case, and special attention should be paid to the case when $k$ is the true known value. For true $k$, minimax linkage does not always perform the best in terms of all the evaluation metrics studied, including maximum minimax radius. 
This paper was motivated by the IFCS Cluster Benchmarking Task Force's call for clustering benchmark studies and the white paper \cite{benchmark}, which put forth guidelines and principles for comprehensive benchmarking in clustering. Our work is designed to be a neutral benchmark study of minimax linkage.
\end{abstract}

%
%

%

\maketitle

\section{Introduction}
\label{sec:intro}

Hierarchical agglomerative clustering involves successively grouping items within a dataset together, based on similarity of the items. The algorithm finishes once all items have been linked, resulting in a hierarchical group similarity structure. Given that two items are merged together, we must determine how similar that merged group is to the remaining items (or groups of items). In other words, we have to recalculate the dissimilarity between any merged points. This dissimilarity between groups can be defined in many ways, and these are known as linkage methods. Standard, established linkage methods include single, complete, average and centroid linkage. Minimax linkage, which was first introduced in \cite{ogMinimax} and formally analyzed in \cite{Bien2011}, will be the subject of our evaluation. We describe hierarchical agglomerative clustering and the linkage methods precisely as follows.

Given items $1, \ldots, n$, dissimilarities $d_{ij}$ between each pair $i$ and $j$, and dissimilarities $d(G, H)$ between groups $G = \{i_1, i_2, \ldots i_r\}$ and $H = \{j_1, j_2, \ldots, j_s\}$, hierarchical agglomerative clustering starts with each node in a single group, and repeatedly merges groups such that $d(G, H)$ is within a threshold $D$. $d(G, H)$ is determined by the linkage method, defined as follows:

\begin{itemize}
	\item[] \textbf{Single linkage} $d_{\text{single}}(G, H) = \min_{i \in G, j \in H} d_{ij}$. The distance between two clusters is defined as the distance between the closest points across the clusters. 
	\item[] \textbf{Complete linkage} $d_{\text{complete}}(G, H) = \max_{i \in G, j \in H} d_{ij}$. The distance between two clusters is defined as the distance between the farthest points across the clusters.  
	\item[] \textbf{Average linkage} $d_{\text{average}}(G, H) = \frac{1}{|G||H|} \sum_{i \in G, j \in H} d_{ij}$. The distance between two clusters is defined as the average of the distances across all pairs of points across the clusters. %
	\item[] \textbf{Centroid linkage} $d_{\text{centroid}}(G, H) = d(\bar{G}, \bar{H})$, where $\bar{G} = \frac{i_1 + i_2 + \hdots + i_r}{r}$ and $\bar{H} = \frac{j_1 + j_2 + \hdots + j_s}{s}$. The distance between two clusters is defined as the distance between the centroid (mean) of the points within the first cluster and the centroid (mean) of the points in the second cluster. Often, the centroids have no intuitive interpretation (e.g., when items are text or images).  
	\item[] \textbf{Minimax linkage} $d_{\text{minimax}}(G, H) = \min_{i \in G\cup H} r(i, G \cup H)$, where $r(i, G) = \max_{j \in G} d_{ij}$, the radius of a group of nodes $G$ around $i$. Informally, each point $i$ belongs to a cluster whose center $c$ satisfies $d_{ic} \leq D$.
\end{itemize}


Bien and Tibshirani \cite{Bien2011} expand upon Ao et al. \cite{ogMinimax} by providing a more comprehensive evaluation of minimax linkage. In particular, they compare minimax linkage to the standard linkage methods using five data sets and two different evaluation metrics. Additionally (although not the focus of the current paper), the authors prove several theoretical properties, for example that dendrograms produced by minimax linkage cannot have inversions and are robust to some data perturbations. They also perform additional evaluations, compare prototypes to centroids, and benchmark computational speed. 

The comparisons of minimax linkage to standard linkage methods in \cite{Bien2011} are summarized in Table \ref{tab:bienEval}. For the colon and prostate cancer data sets, distance to prototype was calculated for minimax linkage, but not for the other linkage methods, since those two data sets were used to compare prototypes to centroids, rather than compare the different linkage methods. More details on the data sets and metrics used are in Sections \ref{sec:benchmark} and \ref{sec:evaluation}.

\begin{table}[!ht]
	\caption {Comparisons to standard linkage methods in \cite{Bien2011}? \label{tab:bienEval} }
	\centering
	\begin{tabular}{llll}
		\hline 
		Dataset & Distance to  & Misclassification  \\
		 &  prototype &  rate  \\
		\hline 
		Olivetti Faces & Yes & No   \\
		\hline 
		Grolier Encyclopedia & Yes & No     \\	
		\hline
		Colon Cancer & Not quite & No      \\	
		\hline
		Prostate Cancer & Not quite & No         \\	
		\hline
		Simulations & No & Yes  \\	
		\hline
	\end{tabular}
\label{tab:bienSum}
\end{table}

``Benchmarking in cluster analysis: A white paper'' \cite{benchmark} makes multiple recommendations for analyses of clustering methods. We focus on those for data sets and evaluation metrics. 

The first recommendation in \cite{benchmark} with respect to choosing data sets is to ``make a suitable choice of data sets and give an explicit justification of the choices made.'' This was not done thoroughly in the original Bien and Tibshirani paper. It was not explained why the particular data sets were chosen for the different evaluations, and features of the data sets were not fully described. In our study, we both add additional data sets and justify existing ones (which include both synthetic and empirical data) in Section \ref{subsec:data}. 

With respect to evaluation metrics, \cite{benchmark} recommends that we think carefully about criteria used and justify our choices. They also recommend that we ``consider multiple criteria if appropriate." Additionally, criteria should be applied across all data sets, and this is one of our main critiques of the existing evaluation, where not all of the data sets used were evaluated on all the criteria suggested (in Table \ref{tab:bienEval}, all cells should be ``Yes''). 

Distance to prototype was well-justified (this is the crux of minimax linkage), but not misclassification rate. While interpretable cluster representatives are important, a researcher may also care about how accurately the algorithm classifies the items in the data set. That being said, when there are a large number of small clusters, the misclassification rate might not be the best measure of performance. In such cases, when working with pairwise comparisons, there is often a large class imbalance problem; most pairs of items do not truly match. A method could achieve a very low misclassification rate simply by predicting all pairs to be non-matches. Therefore we chose to include precision and recall as an additional metric to evaluate clustering quality. 

Finally, a suggestion of \cite{benchmark} is to fully disclose data and code. Unlike the original paper, we supply the code and data that accompanies this paper, for full reproducibility. We have also written an R package, \texttt{clusterTruster}, available on GitHub (\url{https://github.com/xhtai/clusterTruster}), which allows the performance of additional evaluations on user-supplied data.

This paper is designed to be a neutral benchmark study of minimax linkage, and the specific contributions are:

\begin{enumerate}
	\item{An evaluation of all data sets on all of the criteria in \cite{Bien2011}}
	\item{A better assessment of performance with the utilization of precision and recall}
	\item{An evaluation on additional (diverse) data sets not in \cite{Bien2011}}
	\item{Providing publicly available code and an R package that allow for full reproducibility and transparency, while simplifying the process of making additional evaluations on user-supplied data}.
\end{enumerate}

The rest of the paper is organized as follows. Section \ref{sec:benchmark} describes our benchmark study, including justifications for the data sets and evaluation metrics used. Section \ref{sec:evaluation} presents the results, and Section \ref{sec:discussion} concludes.
\section{Benchmark Study}
\label{sec:benchmark}

In this benchmark study we both introduce new evaluation metrics (which we apply to every data set), and add new data sets to provide for a more comprehensive analysis. These are detailed as follows. 

\subsection{Evaluation metrics}
to ``
Our first improvement to \cite{Bien2011} is to utilize all evaluation metrics provided in the paper on all of the data sets (as opposed to some metrics on some data sets). In other words, any instance of ``Not quite'' or ``No'' within Table \ref{tab:bienEval} to should be changed to ``Yes.'' Additionally, we introduce precision and recall as additional evaluation metrics. The evaluation metrics used are described as follows.

\textbf{Distance to prototype} 

The distance to prototype is measured by the maximum minimax radius. The radius of a group of nodes $G$ around $i$ was defined in Section \ref{sec:intro}, as $r(i, G) = \max_{j \in G} d_{ij}$. This is the distance of the farthest point in cluster G from point $i$.

The prototype is selected to be the point in G with the minimum radius, and this radius is known as the minimax radius, 
\begin{equation} \label{eq:minimaxRadius}
r(G) = \min_{i \in G} r(i, G).
\end{equation}

(Using this notation, minimax linkage between two clusters $G$ and $H$ can also be written as $r(G\cup H)$.)

Now, for a clustering with $k$ clusters, each of the $k$ clusters is associated with a minimax radius, $r(G_k)$. We consider the maximum minimax radius, $\max_{k} r(G_k)$, in other words the ``worst'' minimax radius across all clusters. In this sense, the maximum minimax radius can be thought of as a measure of the tightness of clusters around their prototype. A small value indicates that points within the cluster are close to their prototypes, meaning that the prototype is an accurate representation of points within the cluster. 

Minimax linkage relies on successively merging clusters to produce the smallest maximum radius of the resulting cluster, so we would expect minimax linkage to perform the best among other linkage methods in terms of producing the smallest maximum minimax radii.

\textbf{Misclassification rate}

The misclassification rate is defined as the proportion of misclassified examples out of all the examples. 
$$\text{Misclassification rate} = \frac{\text{Number of misclassified examples}}{\text{Total number of examples}}$$

In the clustering context, misclassification rate is defined on pairs of items, specifically we consider each of the $\binom n 2$ pairs, where $n$ is the number of individual items, and the outcome of interest is whether the pair is predicted to be in the same cluster or not. A pair is misclassified if the clustering method predicts that the pair is in the same cluster when the true clustering says they are not, or vice versa.

A low misclassification rate typically indicates high accuracy (a good classifier). But, in cases with a large class imbalance (typically many non-matches and few matches) we need to be careful with using misclassification rates because simply classifying all items as non-matches produces a very low misclassification rate. 

\textbf{Precision and recall}

To take into account class imbalance, we use the evaluation metrics precision and recall. A typical confusion matrix is below.

\begin{center}
\begin{tabular}{llr}
	\multicolumn{3}{r}{Predicted} \\
	\cline{2-3}
	Truth & 0 & 1 \\
	\hline
	0 & TN & FP \\
	1 & FN  & TP \\
	\hline
\end{tabular}
\end{center}

\begin{center}
\begin{minipage}[t]{.49\linewidth}
	\begin{center}
			$$\text{Precision} = \frac{\text{TP}}{\text{TP} + \text{FP}}$$
	\end{center}
\end{minipage}
\begin{minipage}[t]{.49\linewidth}
	\begin{center}
			$$\text{Recall} = \frac{\text{TP}}{\text{TP} + \text{FN}}$$
	\end{center}
\end{minipage}
\end{center}

%

Both precision and recall do not include the true negative cell in their calculation and therefore produce fairer estimates of accuracy in class imbalanced data sets, which are common to clustering. The maximum value for precision and recall are both 1, and a good classifier should have high precision and recall. 

\textbf{All $k$, best $k$ vs true $k$}

Again, define $k$ as the number of clusters in the clustering. In \cite{Bien2011}, evaluation for distance from prototype was conducted over all possible values of $k$ (specifically in a data set of $n$ items, $k \in [1, n] $). Misclassification rate however was reported for the best $k$, meaning the lowest misclassification rate over all $k$, and the true $k$, where the ground truth clustering is known.

In this paper we evaluate on all metrics using all $k$, and also report the metrics for true $k$. It is possible to derive measures for the best $k$, but due to the large number of data sets and evaluation metrics used, this became somewhat intractable and was not pursued further, but can be a subject of future work.


\subsection{Data sets} \label{subsec:data}
In terms of the data sets considered, we use all of the data used in \cite{Bien2011} (except for Grolier Encyclopedia), and introduce additional data sets that exhibit a wider range of data attributes. These additional data sets were included also to ensure that those used in \cite{Bien2011} were not deliberately selected to produce desired results. The Grolier Encyclopedia data set does not include true clusters and was therefore not included in the current paper. Brief descriptions are as follows, and more details for many of the data sets can be found in \cite{Bien2011}. A summary of the data is in Table \ref{tab:allData}. 

\textbf{Olivetti Faces}
This data contains 400 images of 64 $\times$ 64 pixels. There are 10 images each from 40 people. The pairwise distance measure used is $l_2$ distance. Here we use the data from the \texttt{RnavGraphImageData} package in R. 

\textbf{Colon Cancer}
The Colon Cancer data set contains gene expression levels for 1000 genes for 62 patients, 40 with cancer and 22 healthy. The pairwise distance measure used is correlation. Here we use the data from the \texttt{HiDimDA} package in R. 

\textbf{Prostate Cancer}
The Prostate Cancer data contains gene expression levels for 6033 genes for 102 patients, 52 with cancer and 50 healthy. The pairwise distance measure used is correlation. There are multiple versions of the data available online and in R packages. The version we use is from \url{https://stat.ethz.ch/~dettling/bagboost.html}, and our results match the resulting plots produced in \cite{Bien2011}.  

\textbf{Simulations}
We repeat the simulations done in \cite{Bien2011}. These involve three sets of data: spherical, elliptical and outliers. Each data set has 3 clusters of 100 points each in $\mathbb{R}^{10}$. Both $l_1$ and $l_2$ distances are used as pairwise distance measures. In \cite{Bien2011} simulations were run 50 times each, but here we only ran each once. In future analyses it is possible to perform more runs. 

\textbf{Iris}
The iris data set \cite{Anderson1936,Fisher1936} is pre-loaded in R and has been used extensively as an example data set in various applications, including clustering. It contains 50 flowers from each of 3 species. There are four features for each observation, sepal length and width and petal length and width. Here we simply scale and center the features and use $l_2$ distance as a pairwise distance measure. 

\textbf{NBIDE and FBI S\&W}
The National Institute of Standards and Technology (NIST) maintains the Ballistics Toolmark Research Database (\url{https://tsapps.nist.gov/NRBTD}), containing images of cartridge cases from test fires of various firearms. These are of 3D topographies, meaning that surface depth is recorded at each pixel location. Each image is approximately 1200 $\times$ 1200 pixels. We use data from two different data sets, NIST Ballistics Imaging Database Evaluation (NBIDE) \cite{Vorburger2007} and FBI Smith \& Wesson. The former contains 12 images each from 12 different firearms, and the latter contains 2 images each from 69 different firearms. We have pre-processed and aligned these images using the R package \texttt{cartridges3D} (available at \url{https://github.com/xhtai/cartridges3D}), and extracted a correlation between each pair of images. The resulting pairwise comparison data are available in the \texttt{clusterTruster} package. 

\begin{table}[!ht]
	\caption {Data sets \label{tab:allData} }
	\centering
	\begin{tabular}{lll}
		\hline 
		 Data set & Included & Description \\
		& in \cite{Bien2011}?  &  \\
		\hline 
		\pbox{3cm}{\textbf{Olivetti Faces}  \\
		(Roweis) } & Yes &  \pbox{4.1cm}{ n = 400, p = 4096 \\
			$k$ = 40  \\
			image data of human faces} \\
		\hline
		\pbox{3cm}{\textbf{Colon Cancer} \\
		(Alon et al. 1999) } & Yes & \pbox{4.1cm}{ n = 62, p = 2000 \\
			 $k$ =  2 \\
			 high dimensional data} \\	
		\hline
		\pbox{3cm}{\textbf{Prostate Cancer}  \\
			(Singh et al. 2002)} & Yes & \pbox{4.1cm}{ n = 102, p = 6033 \\
		     $k$ =  2 \\
		     high dimensional data } \\	
		\hline
		\pbox{3cm}{\textbf{Spherical}  \\
		(Bien et al. 2011) } & Yes & \pbox[c]{4.1cm}{ n = 300, p = 10 \\
			$k$ = 3 \\
			spherical shape\\
			 L-1 and L-2 distance used} \\	\hline
		\pbox{3cm}{\textbf{Elliptical}  \\
		(Bien et al. 2011) }& Yes &  \pbox[c]{4.1cm}{ n = 300, p = 10 \\
			 $k$ = 3 \\
			 elliptical shape\\
			 L-1 and L-2 distance used} \\	\hline
		\pbox{3cm}{\textbf{Outlier}  \\
		(Bien et al. 2011) }& Yes &  \pbox[c]{4.1cm}{ n = 300,  p = 10 \\
			 $k$ = 3 \\
			  spherical shape with outliers \\
			   L-1 and L-2 distance used} \\	
		\hline
		\pbox{3cm}{\textbf{Iris}  \\
			(Anderson, 1936; Fisher, 1936) }& No &  \pbox[c]{4.1cm}{ n = 150,  p = 4 \\
			$k$ =  3 \\
			elliptical shape \\
			well-separated clusters} \\	
		\hline
		\pbox{3cm}{\textbf{NBIDE}  \\
			(Vorburger et al. 2007) }&  No &  \pbox[c]{4.1cm}{ n = 144,  p = 144,000 \\
			$k$ = 12 \\
			image data of cartridge cases } \\	
		\hline
		\pbox{3cm}{\textbf{FBI S\&W}  \\
			 }& No &  \pbox[c]{4.1cm}{ n = 138,  p = 144,000 \\
			$k$ =  69 \\
			large number of small clusters } \\	
		\hline
	\end{tabular}
\end{table}


\section{Evaluation Results}
\label{sec:evaluation}

We report our evaluation results for all data sets and evaluation metrics, using all $k$ and true $k$. In Tables \ref{tab:results} and \ref{tab:results2}, we report the values for the maximum minimax radius, misclassification rate, precision and recall for all linkage types (single, complete, average, centroid, and minimax linkage) for the case where $k$ = true $k$ for each data set. In Figures \ref{fig:faces} through \ref{fig:NBIDE}, we show the distribution of maximum minimax radius, misclassification, and precision-recall across all possible values of $k$.

\subsection{Results for true $k$ \label{sec:bestk}}

It is important to understand how our evaluation metrics change for multiple values of $k$, especially because $k$ is often unknown. That being said, it is common to know a plausible range of $k$ values and therefore results in out-of-scope regions may be irrelevant. We present the clustering results for the true value of $k$ for each data set in Tables \ref{tab:results} and \ref{tab:results2} in the Appendix. As an example, we reproduce the results for the Olivetti Faces data set in Table \ref{tab:faces}.

\begin{table}[h]
\caption{Results for Olivetti Faces data set with $k = 40$ (true $k$)}
\begin{tabular}{lllll}
	\hline 
	Linkage type & Max minimax & Misclass- & Precision & Recall \\
	 					 & radius & ification  &  &  \\
	\hline 
	single & 3394.93 & 0.40 & 0.04 & 0.78 \\ 
	complete & 2606.25 & \textbf{0.04} & \textbf{0.31} & 0.49 \\ 
	average & 2449.69 & 0.07 & 0.18 & 0.60 \\ 
	centroid & 3259.74 & 0.79 & 0.02 & \textbf{0.83} \\ 
	minimax & \textbf{2293.45} & 0.05 & 0.24 & 0.57 \\ 
\end{tabular}
\label{tab:faces}
\end{table}

In Table \ref{tab:faces}, we find that hierarchical clustering with minimax linkage produces the smallest maximum minimax radius, indicating that the images within each cluster are close to the cluster's prototype. This means that the prototype is a good representation of the cluster. Using a prototype is especially useful for interpreting cluster results in cases when averages of the data do not make practical sense (e.g., images, text). Minimax linkage does not produce the lowest misclassification rate, although the rate is comparable to both average and centroid linkage. Because average and centroid linkage result in uninterpretable cluster ``representatives,'' minimax linkage could holistically be considered the best performer. However, minimax linkage does not produce the highest precision and recall which we argue in Section \ref{sec:benchmark} should also be reported when determining linkage quality.

In Tables \ref{tab:results} and \ref{tab:results2}, we examine maximum minimax radius (for true $k$) and find that minimax linkage does not always produce the best results. In the Colon Cancer, Prostate Cancer, Spherical-L2, Sperical-L1 and Elliptical-L2 data sets, other linkage methods produce the smallest maximum minimax radius. We also note that the maximum minimax radius produced by minimax linkage is often (but not always) close to the radius of another linkage method. Bien and Tibshirani \cite{Bien2011}  claim that ``minimax linkage indeed does consistently better than the other methods in producing clusterings in which every point is close to a prototype,'' which we do see across multiple $k$ values (Figures \ref{fig:faces} through \ref{fig:NBIDE}; more details in Section \ref{ssec:allK}). However, when we look specifically at the true $k$ case, which is more relevant in practice, our results dispute this claim. 

In terms of the misclassification rate, minimax linkage performs well. In the Elliptical-L2, Spherical-L2, Colon Cancer, and Olivetti Faces datasets, other linkage methods produce smaller misclassification rates, but in each case the minimax linkage rate was close to (within 0.03 of) the best rate. This finding is consistent with the claims in \cite{Bien2011}. 

In terms of precision, minimax linakge performs worse than other linkage methods in the Olivetti Faces, Colon Cancer, Prostate Cancer, Elliptical-L2 and FBISW data sets. For recall, minimax linakge performs worse than other methods in all data sets except NBIDE. 

In summary, we find that for true $k$, minimax linkage does not consistently perform best in terms of smallest maximum minimax radius, highest precision, or highest recall rates. It does consistently perform well in terms of misclassification. One of the core claims in \cite{Bien2011} is that minimax linakge consistently performs best in terms of producing low maximum minimax radius, but we find that it is not consistently the case for the true $k$ scenario.

\subsection{Results across all $k$} \label{ssec:allK}

In this section, we look at how the performance metrics change across different values of $k$, as opposed to the true value of $k$ in Section \ref{sec:bestk}. This could still be relevant in practice where $k$ is unknown, or if we want an overall sense of the performance of the method across all possible clusterings.

\begin{figure}[h]
	\includegraphics[width=\linewidth]{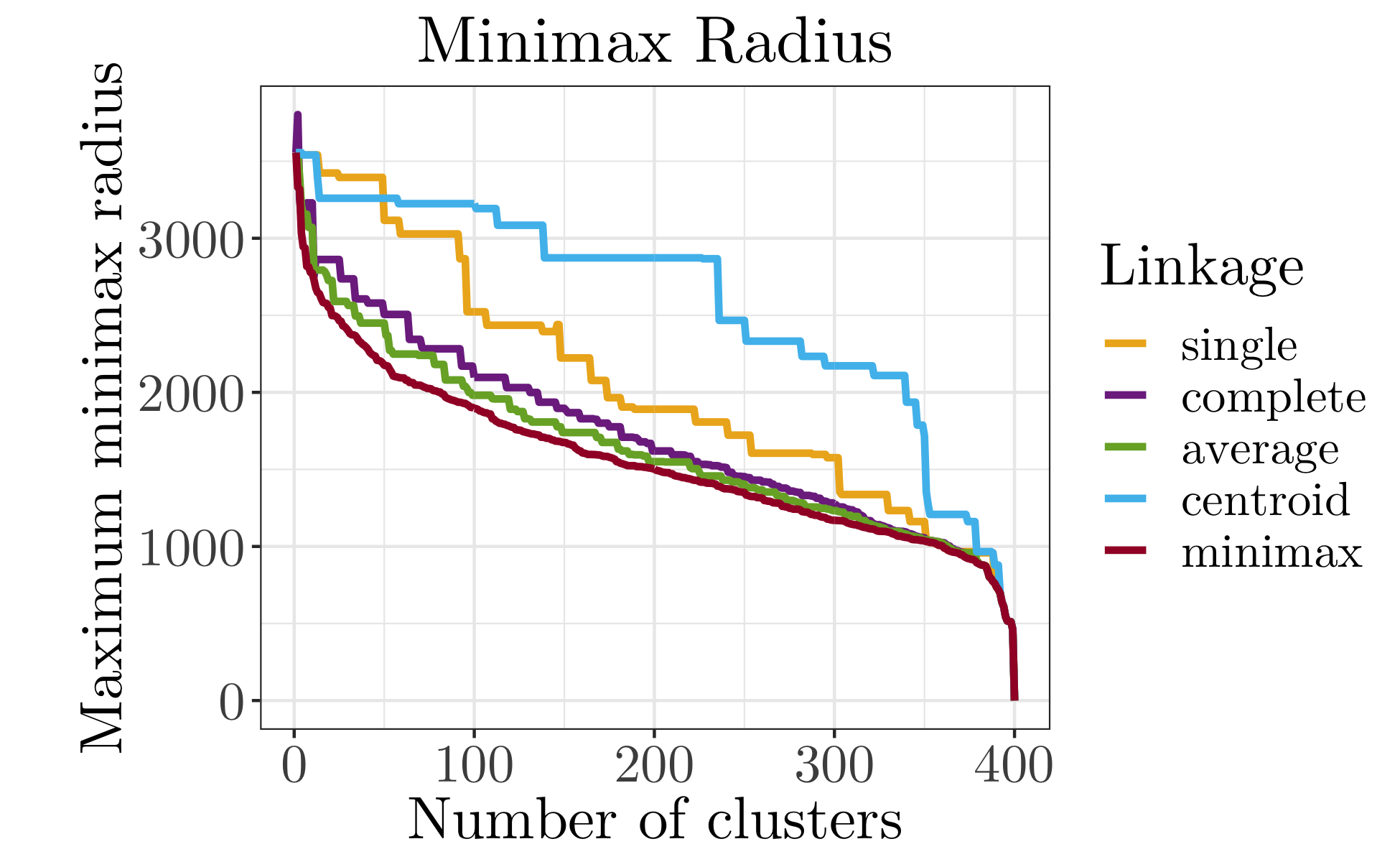}
	\caption{Minimax radius results for Olivetti Faces}
	\label{fig:facesMinimax}
\end{figure}

\begin{figure}[h]
	\includegraphics[width=\linewidth]{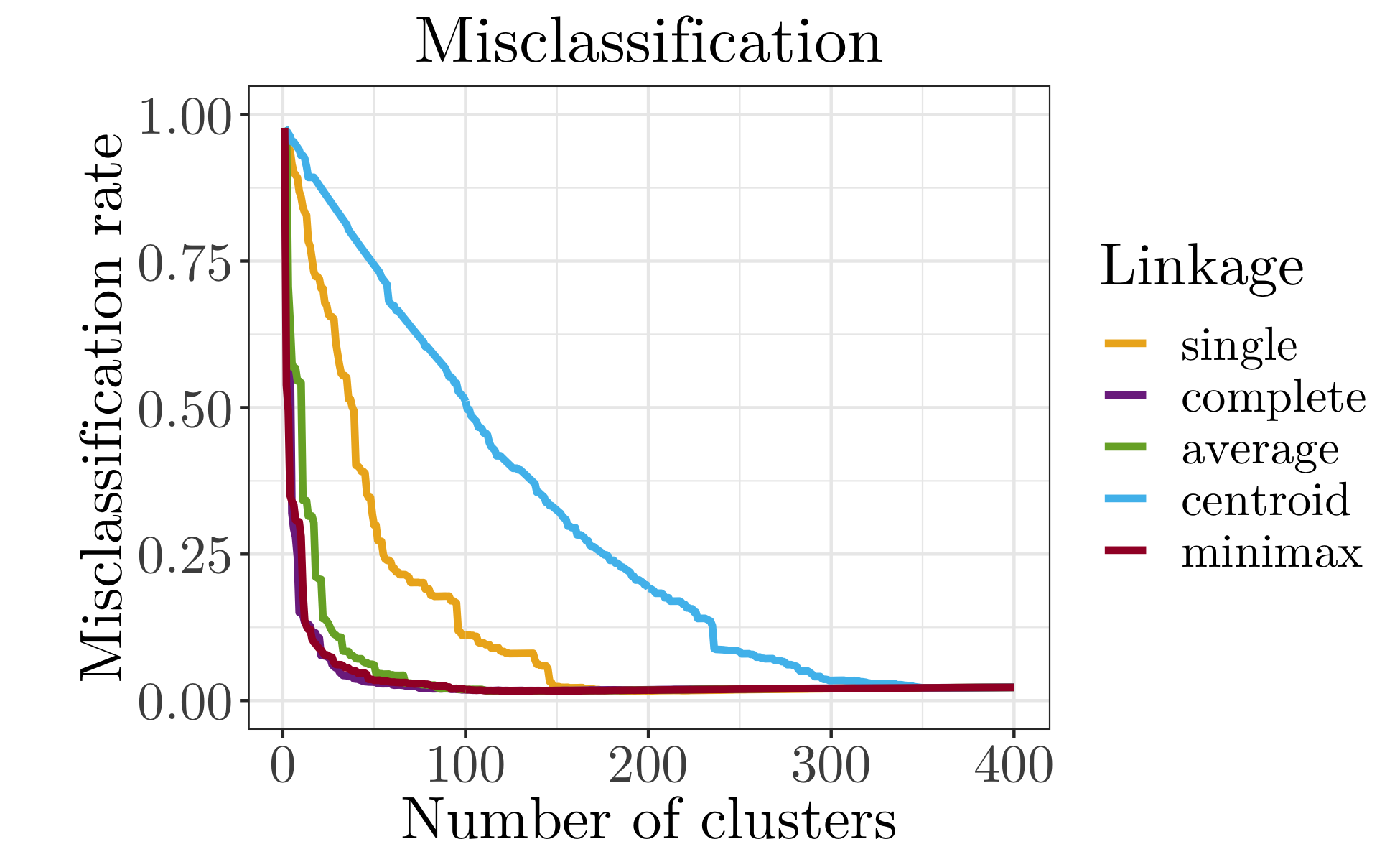}
	\caption{Misclassification results for Olivetti Faces}
	\label{fig:facesMisclass}
\end{figure}

\begin{figure}[h]
	\includegraphics[width=\linewidth]{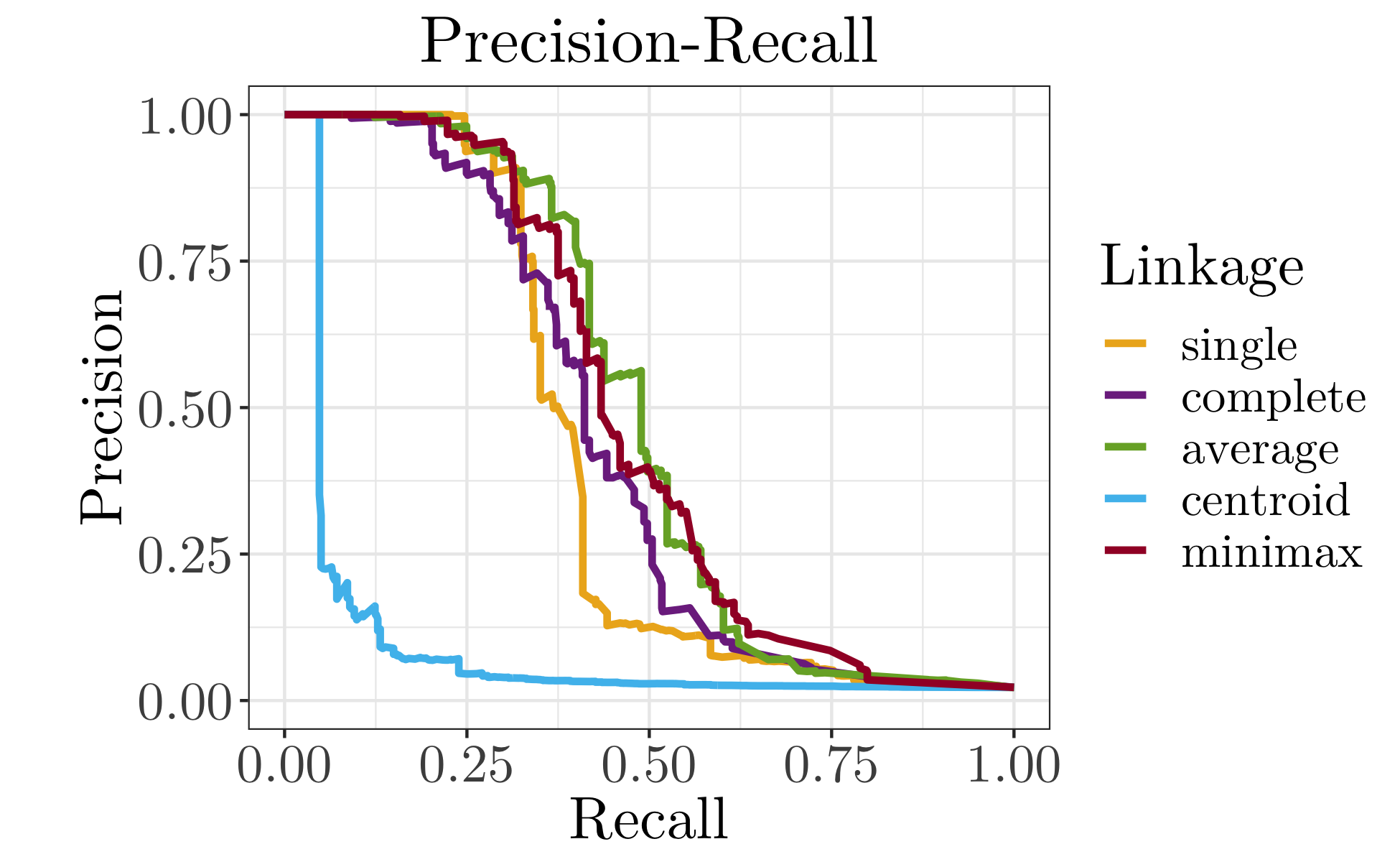}
	\caption{Precision vs. Recall results for Olivetti Faces}
	\label{fig:facesPr}
\end{figure}

The full results are in Figures \ref{fig:faces} through \ref{fig:NBIDE} in the Appendix (Section \ref{sec:appendix}). Here we again use the Olivetti Faces data set to illustrate the results. We found that for true $k$ (40), minimax linkage performed best in terms of maximum minimax radius, but not in terms of misclassification, precision, or recall. The results for all $k$ are in Figures \ref{fig:facesMinimax} through \ref{fig:facesPr}.

In Figure \ref{fig:facesMinimax} we find that the line for minimax linkage (smooth dark red line) is consistently lower than the other linkage methods (single, complete, average, and centroid).  

In Figure \ref{fig:facesMisclass}, we see that minimax linkage (smooth dark red line) consistently performs similarly to complete linkage (smooth dark purple line) and both methods do better than single and centroid linkage. Average linkage performs similarly to complete and minimax linkage for large values of $k$.  

In Figure \ref{fig:facesPr} we plot the precision versus recall graph. For each value of $k$ we plot both precision and recall, and connect the values in order of increasing $k$. The area under the curve has maximum value 1, and we want this to be as large as possible. In the Olivetti Faces data set in Figure \ref{fig:facesPr}, we see that centroid linkage performs poorly for most values of $k$. Average linkage appears to perform best for most values of $k$ although minimax and single linkage also perform well for certain values of $k$.  

As mentioned, the full results for all data sets are in Figures \ref{fig:faces} through \ref{fig:NBIDE} in the Appendix (Section \ref{sec:appendix}). Across all data sets, hierarchical clustering with minimax linkage performs best across most values of $k$ in terms of lowest maximum minimax radius. No linkage type stands out as the best for misclassification performance across the different data sets. Similarly for precision and recall, there is no best performing linkage method.
\section{Discussion and Conclusion}
\label{sec:discussion}

Bien and Tibshirani analyzed minimax linkage in 2011 \cite{Bien2011}, and performed comparisons to standard linkage methods. We expand on this evaluation, taking into account guidelines recommended in ``Benchmarking in cluster analysis: A white paper'' \cite{benchmark}: we justify choices of data sets and evaluation metrics, apply criteria across all data sets, and fully disclose data and code. Comparing to \cite{Bien2011}, we use additional data sets, include precision and recall as additional evaluation metrics, and use all metrics for all data sets. We evaluate on all possible clusterings $k$, as well as the true number of clusterings. We highlight results for the latter case, since metrics for this value of $k$ are often more relevant than results across all values of $k$. 

One of the main claims of \cite{Bien2011} is that minimax linakge consistently performs best in terms of producing results with low maximum minimax radius, but we find that it is not consistently the case for the true $k$ scenario. We do find (similarly to \cite{Bien2011}), that minimax linkage often produces the smallest maximum minimax radius (compared to other linkage methods) across all possible values of $k$. This means that overall, minimax linkage does produce clusters where objects in a cluster are tightly clustered around their prototype. Prototypes are a good representation of their cluster and have good interpretability. We find that minimax linkage performs well in terms of misclassification across all data sets, but it does not always produce high precision and recall (which we suggest should also be reported due to the common class imbalance problem in cluster analysis). 

In our comprehensive analysis of minimax linakge we came to two main conclusions. 

\begin{enumerate}
	\item \textbf{For true $k$}: Minimax linkage does not consistently perform best in terms of smallest maximum minimax radius, highest precision, or highest recall. It does consistently perform well in terms of misclassification.
	\item \textbf{Across all $k$}: Minimax linkage performs best across most values of $k$ in terms of lowest maximum minimax radius. No linkage type stands out as the best for misclassification performance, precision, or recall across all data sets.
\end{enumerate}

Future work will include an increased focus on simulations. The priority of this paper was evaluating performance on real clustering applications, and we included one run of the simulations that were done in \cite{Bien2011}. More work will be done on this in the future, to properly quantify standard errors associated with the metrics. We also noted that it is possible to derive and report measures for the best $k$ as opposed to true $k$, but due to the large number of data sets and evaluation metrics used, this was not pursued but left for future work. Another question that is of interest but was out of the scope of this paper is to analyze cases where the true number of clusters is not known. For example, we might be interested in whether it is easier to discover the number of true clusters using minimax linkage as opposed to other methods, and this is an interesting question that to our knowledge has not been explored. Finally, one issue that was evaluated briefly in \cite{Bien2011}, but that we have not focused on, was computational complexity. This will be the subject of future work.

\bibliographystyle{ACM-Reference-Format}
\bibliography{local,xtai,kfrisoli}

\newpage
\newpage
\section{Appendix}
\label{sec:appendix}

\begin{table*}[!ht]
	\caption {Results for true $k$ \label{tab:results} }
	\centering
	\begin{tabular}{p{3cm} lllll}
		\hline 
		Data set ($k$ = truth) & Linkage type & Max minimax radius & Misclassification & Precision & Recall \\
		\hline 
		\pbox{3cm}{\textbf{Olivetti Faces}  \\
			$k$ = 40 } & single & 3394.93 & 0.40 & 0.04 & 0.78 \\ 
		& complete & 2606.25 & \textbf{0.04} & \textbf{0.31} & 0.49 \\ 
		& average & 2449.69 & 0.07 & 0.18 & 0.60 \\ 
		& centroid & 3259.74 & 0.79 & 0.02 & \textbf{0.83} \\ 
		& minimax & \textbf{2293.45} & 0.05 & 0.24 & 0.57 \\ 
		\hline
	\pbox{3cm}{\textbf{Colon Cancer} \\
		$k$ = 2} & single & 0.34 & \textbf{0.46} & \textbf{0.54} & \textbf{0.98} \\ 
		& complete & \textbf{0.28} & 0.48 & 0.53 & 0.87 \\ 
		& average & \textbf{0.28} & 0.48 & 0.53 & 0.87 \\ 
		& centroid & \textbf{0.28} & 0.47 & 0.53 & 0.90 \\ 
		& minimax & 0.29 & 0.48 & 0.53 & 0.92 \\ 
		\hline
	\pbox{3cm}{\textbf{Prostate Cancer}  \\
		$k$ = 2} & single & 0.48 & 0.50 & 0.50 & \textbf{0.98} \\ 
	& complete & \textbf{0.33 } & 0.49 & 0.50 & 0.77 \\ 
	& average & 0.35 & \textbf{0.49 }& 0.50 & 0.73 \\ 
	& centroid & 0.40 & \textbf{0.49} & 0.50 & 0.69 \\ 
	& minimax & 0.35 & \textbf{0.49} & 0.50 & 0.76 \\ 
		\hline
		\pbox{3cm}{\textbf{Spherical-L2} \\
			$k$ = 3}  &single & 6.07 & 0.66 & 0.33 & \textbf{0.99} \\ 
		& complete & \textbf{5.13} & \textbf{0.24 }& \textbf{0.63} & 0.64 \\ 
		& average & 5.95 & 0.66 & 0.33 & 0.98 \\ 
		& centroid & 6.07 & 0.66 & 0.33 & \textbf{0.99} \\ 
		& minimax & 5.35 & 0.25 & 0.62 & 0.65 \\ 
		\hline
		\pbox{3cm}{\textbf{Spherical-L1} \\
			$k$ = 3}  & single & 15.97 & 0.66 & 0.33 & \textbf{0.99} \\ 
		& complete & \textbf{14.26} & \textbf{0.33} & \textbf{0.51 }& 0.55 \\ 
		& average & 15.72 & 0.66 & 0.33 & 0.98 \\ 
		& centroid & 15.75 & 0.66 & 0.33 & \textbf{0.99 }\\ 
		& minimax & 14.87 & \textbf{0.33} & \textbf{0.51} & 0.51 \\ 
		\hline
		\pbox{3cm}{\textbf{Elliptical-L2} \\
			$k$ = 3}  & single & 6.79 & 0.66 & 0.33 & \textbf{0.99} \\ 
		& complete & \textbf{5.95} & \textbf{0.35} & \textbf{0.48} & 0.51 \\ 
		& average & 6.66 & 0.66 & 0.33 & 0.96 \\ 
		& centroid & 6.76 & 0.66 & 0.33 & \textbf{0.99} \\ 
		& minimax & 6.21 & 0.38 & 0.44 & 0.51 \\ 
		\hline
		\pbox{3cm}{\textbf{Elliptical-L1} \\
			$k$ = 3}  & single & 17.40 & 0.66 & 0.33 & \textbf{0.99} \\ 
		& complete & 16.40 & 0.38 & 0.44 & 0.59 \\ 
		& average & 17.40 & 0.66 & 0.33 & 0.96 \\ 
		& centroid & 17.37 & 0.66 & 0.33 & \textbf{0.99} \\ 
		& minimax & \textbf{15.60} & \textbf{0.33} & \textbf{0.50} & 0.57 \\ 
		\hline
		\pbox{3cm}{\textbf{Outliers-L2} \\
			$k$ = 3}  & single & 6.46 & 0.66 & 0.33 & \textbf{0.99} \\ 
		& complete & \textbf{5.81} & 0.46 & 0.38 & 0.65 \\ 
		& average & 6.12 & 0.65 & 0.33 & 0.95 \\ 
		& centroid & 6.37 & 0.66 & 0.33 & 0.98 \\ 
		& minimax & 5.95 & \textbf{0.39} & \textbf{0.44} & 0.65 \\ 
		\hline
		\pbox{3cm}{\textbf{Outliers-L1} \\
			$k$ = 3}  & single & 17.39 & 0.66 & 0.33 & \textbf{0.99} \\ 
		& complete & 15.99 & 0.42 & 0.39 & 0.50 \\ 
		& average & 16.37 & 0.66 & 0.33 & 0.97 \\ 
		& centroid & 16.37 & 0.66 & 0.33 & 0.98 \\ 
		& minimax & \textbf{14.79} & \textbf{0.26} & \textbf{0.60} & 0.61 \\ 
		\hline
	\end{tabular}
\end{table*}

\begin{table*}[!ht]
	\caption {Results for true $k$ (cont.) \label{tab:results2} }
	\centering
	\begin{tabular}{p{3cm} lllll}
		\hline 
		Data set & Linkage type & Max minimax radius & Misclassification & Precision & Recall \\
		\hline 
		\pbox{3cm}{\textbf{Iris} \\
			$k$ = 3}  & single & 2.97 & 0.23 & 0.59 & \textbf{0.99} \\ 
		& complete & 2.19 & 0.20 & 0.67 & 0.79 \\ 
		& average & 2.56 & 0.22 & 0.60 & 0.96 \\ 
		& centroid & 2.97 & 0.23 & 0.59 & \textbf{0.99} \\ 
		& minimax & \textbf{2.09} & \textbf{0.17} & \textbf{0.71} & 0.79 \\ 
		\hline
		\pbox{3cm}{\textbf{NBIDE} \\
			$k$ = 12}  & single & 0.82 & 0.23 & 0.23 & 0.89 \\ 
		& complete & 0.77 & 0.05 & 0.66 & 0.79 \\ 
		& average & 0.75 & 0.03 & 0.77 & 0.91 \\ 
		& centroid & 0.83 & 0.80 & 0.08 & 0.87 \\ 
		& minimax & \textbf{0.73} & \textbf{0.02} & \textbf{0.84} & \textbf{0.92} \\ 
		\hline
	\pbox{3cm}{\textbf{FBISW} \\
			$k$ = 69}  & single & 0.75 & 0.01 & 0.33 & 0.86 \\ 
		& complete & 0.65 & \textbf{0.00} & \textbf{0.83} & \textbf{0.93} \\ 
		& average & 0.63 & \textbf{0.00} & 0.77 & 0.91 \\ 
		& centroid & 0.82 & 0.17 & 0.02 & 0.58 \\ 
		& minimax & \textbf{0.59} & \textbf{0.00} & 0.70 & 0.90 \\ 
		\hline
	\end{tabular}
\end{table*}

\begin{figure*}[!ht]
	\caption{\label{fig:faces}
		Results for Olivetti faces}
	\centering
	\includegraphics[width=\linewidth]{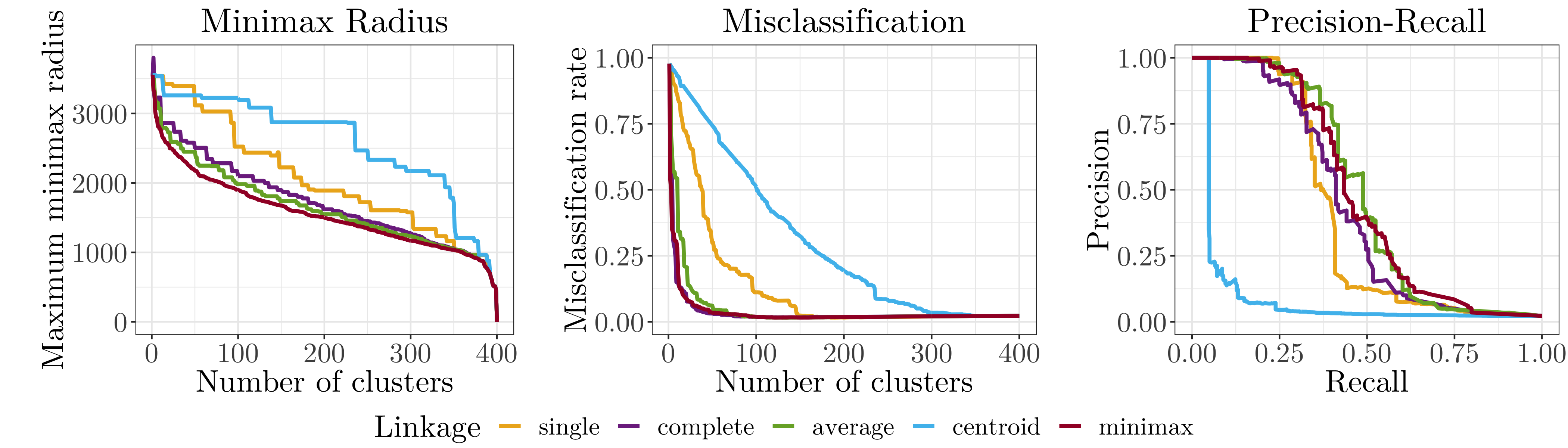}
\end{figure*}

\begin{figure*}[!ht]
	\caption{\label{fig:AlonDS}
		Results for Colon Cancer}
		\centering
		\includegraphics[width=\linewidth]{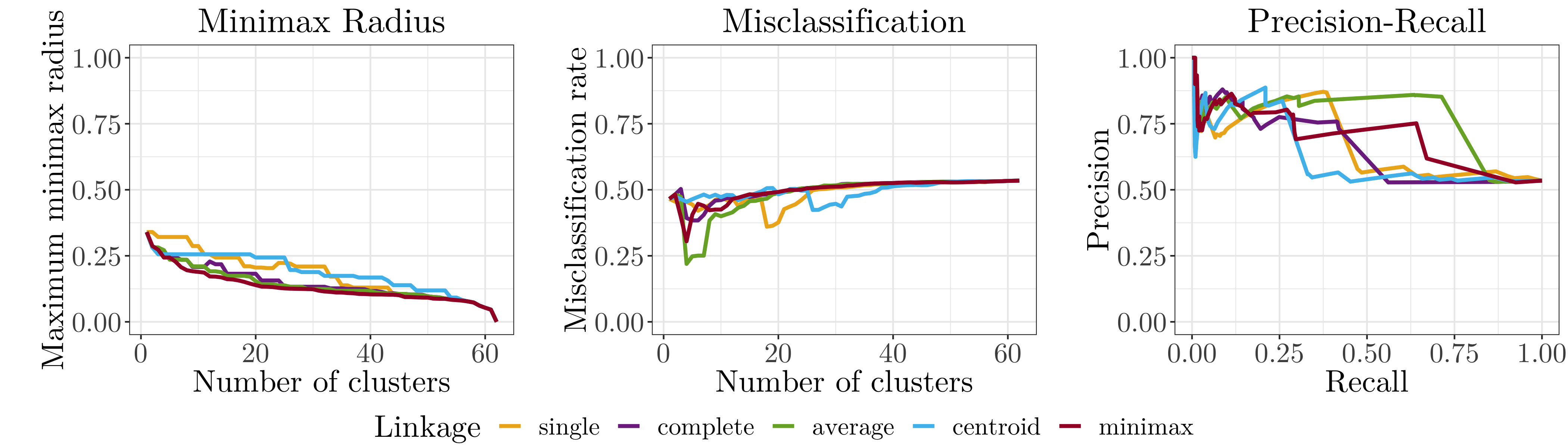}
\end{figure*}

\begin{figure*}[!ht]
	\caption{\label{fig:singh2002}
		Results for Prostate Cancer}
	\centering
	\includegraphics[width=\linewidth]{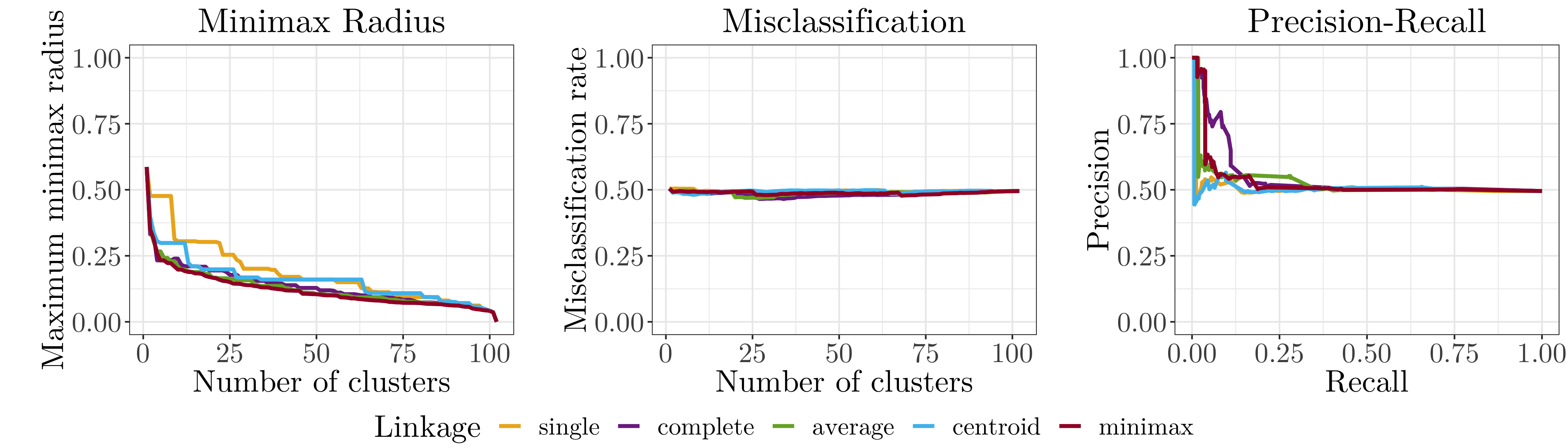}
\end{figure*}

\begin{figure*}[!ht]
	\caption{\label{fig:sphericall2}
		Results for simulation: spherical l2}
	\centering
	\includegraphics[width=\linewidth]{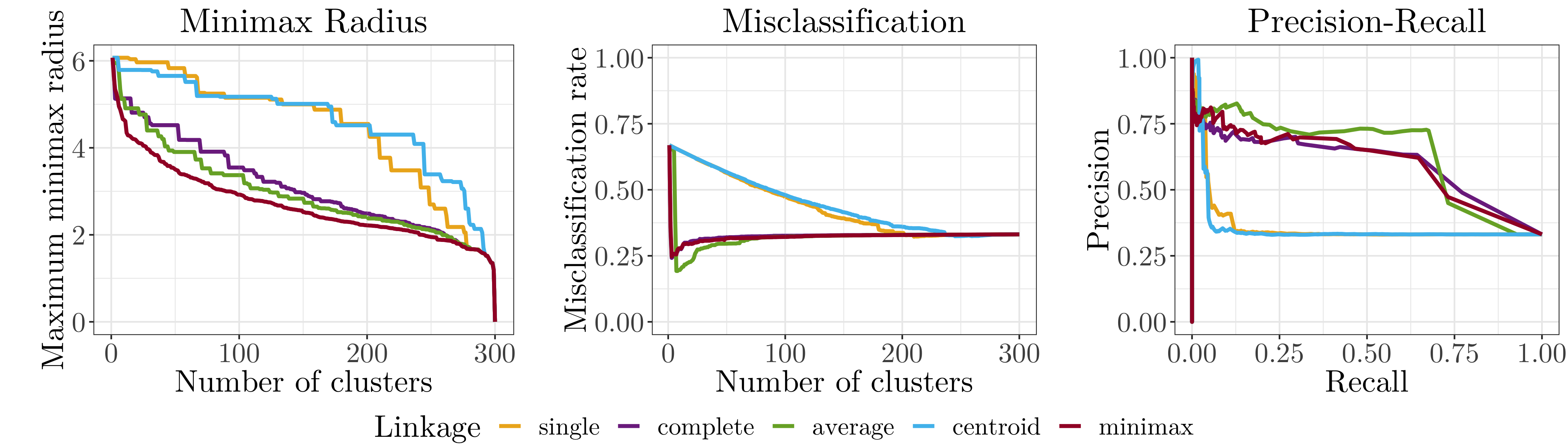}
\end{figure*}

\begin{figure*}[!ht]
	\caption{\label{fig:sphericall1}
		Results for simulation: spherical l1}
	\centering
	\includegraphics[width=\linewidth]{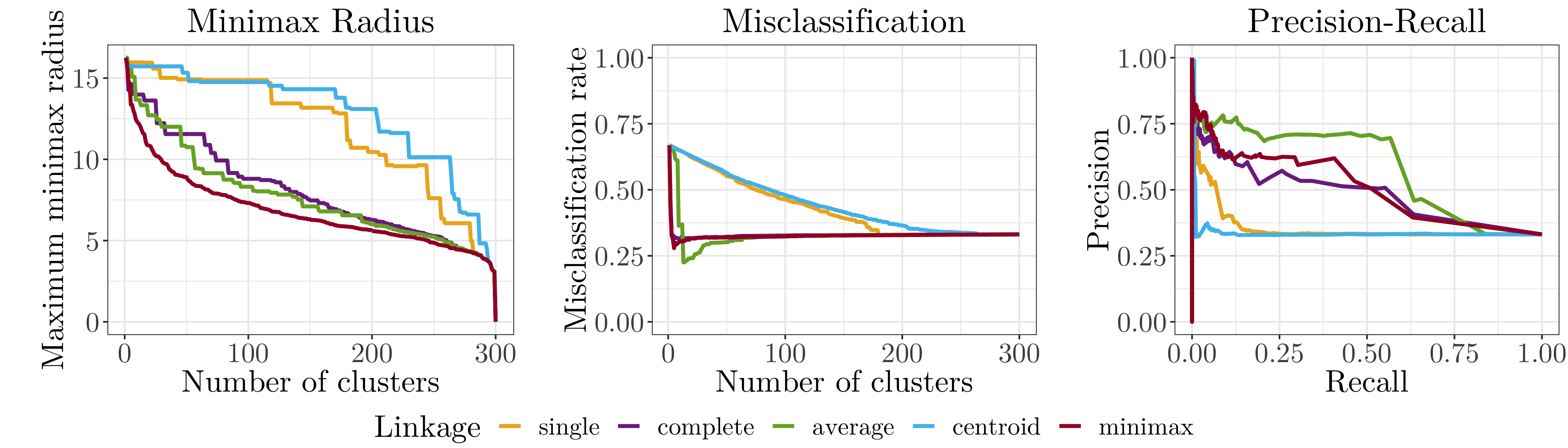}
	
\end{figure*}

\begin{figure*}[!ht]
	\caption{\label{fig:ellipticall2}
		Results for simulation: elliptical l2}
	\centering
		\includegraphics[width=\linewidth]{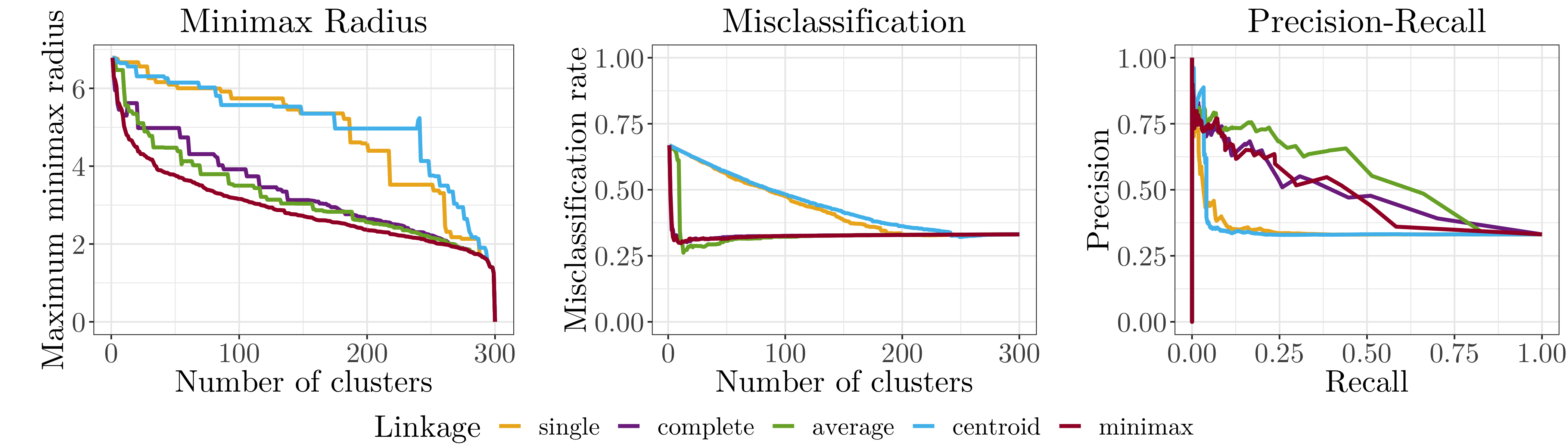}
\end{figure*}

\begin{figure*}[!ht]
	\caption{\label{fig:ellipticall1}
		Results for simulation: elliptical l1}
	\centering
		\includegraphics[width=\linewidth]{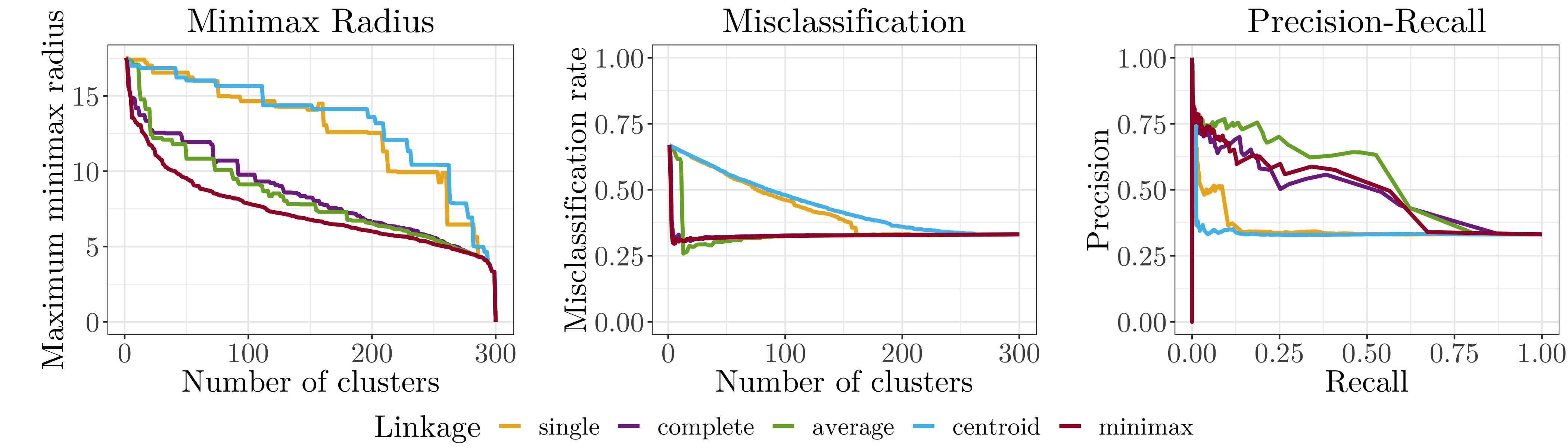}
\end{figure*}

\begin{figure*}[!ht]
	\caption{\label{fig:outliersl2}
		Results for simulation: outliers l2}
	
	\centering
	\includegraphics[width=\linewidth]{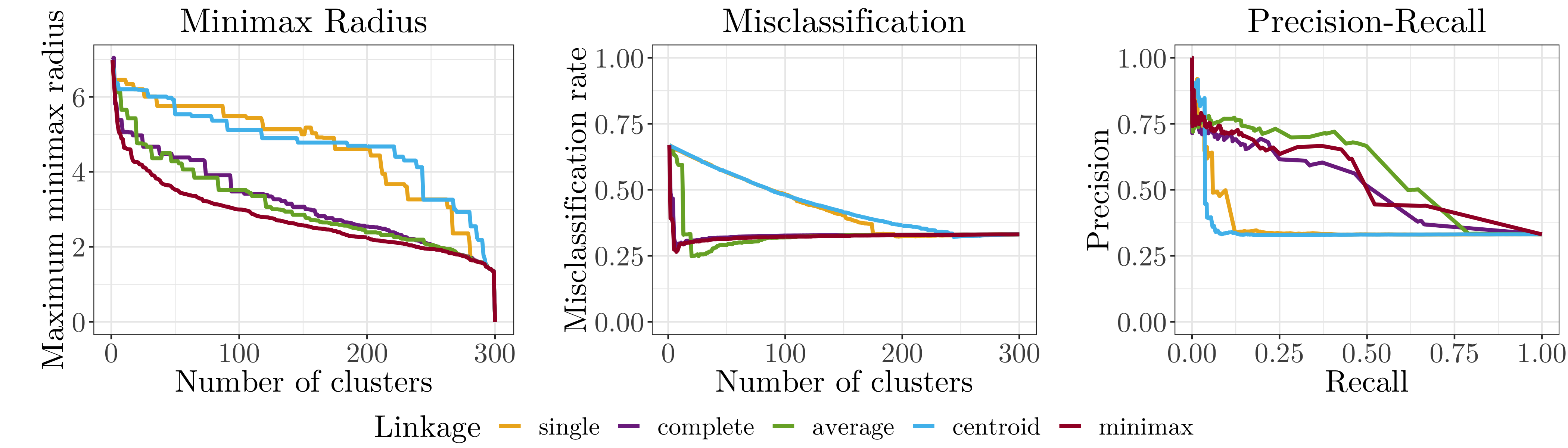}
\end{figure*}

\begin{figure*}[!ht]
	\caption{\label{fig:outliersl1}
		Results for simulation: outliers l1}
	\centering
		\includegraphics[width=\linewidth]{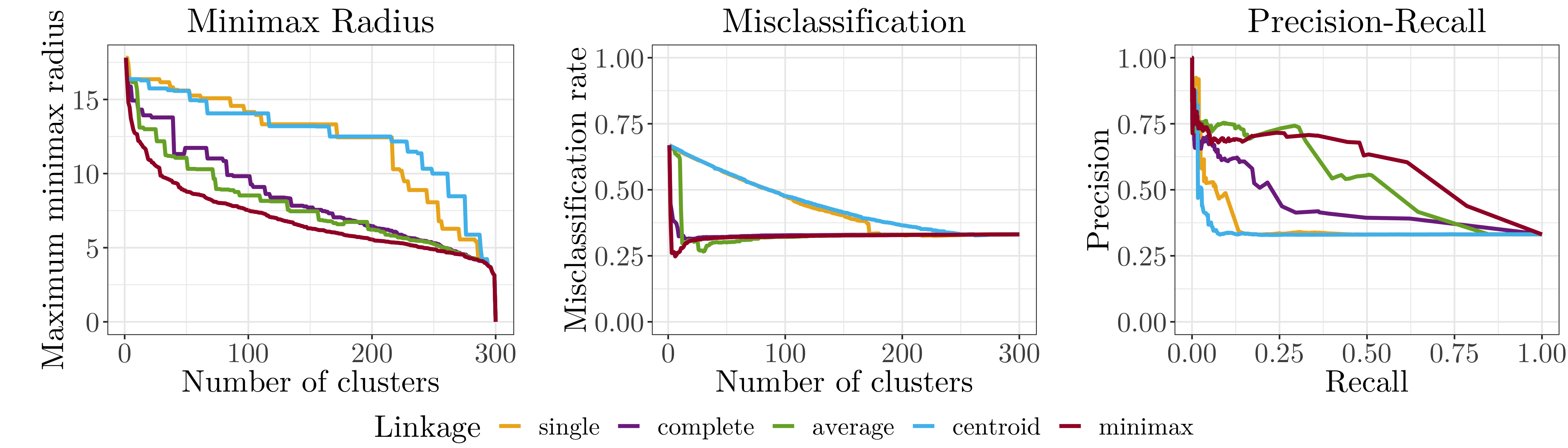}
\end{figure*}

\begin{figure*}[!ht]
	\caption{\label{fig:iris}
		Results for iris}
	\centering
		\includegraphics[width=\linewidth]{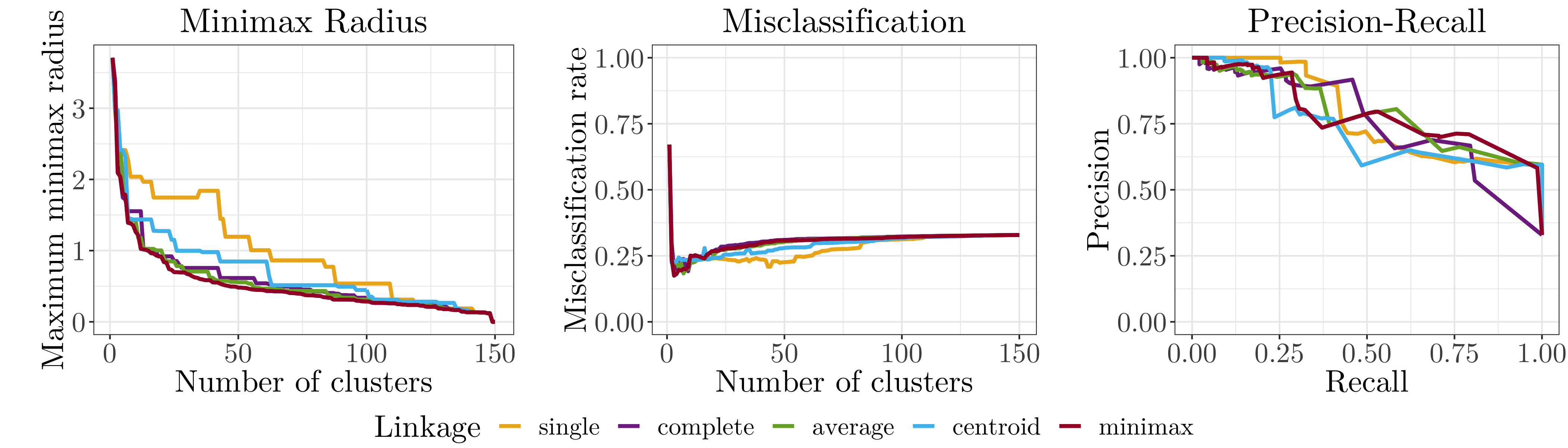}
\end{figure*}

\begin{figure*}[!ht]
	\caption{\label{fig:NBIDE}
		Results for NBIDE study}
	\centering
		\includegraphics[width=\linewidth]{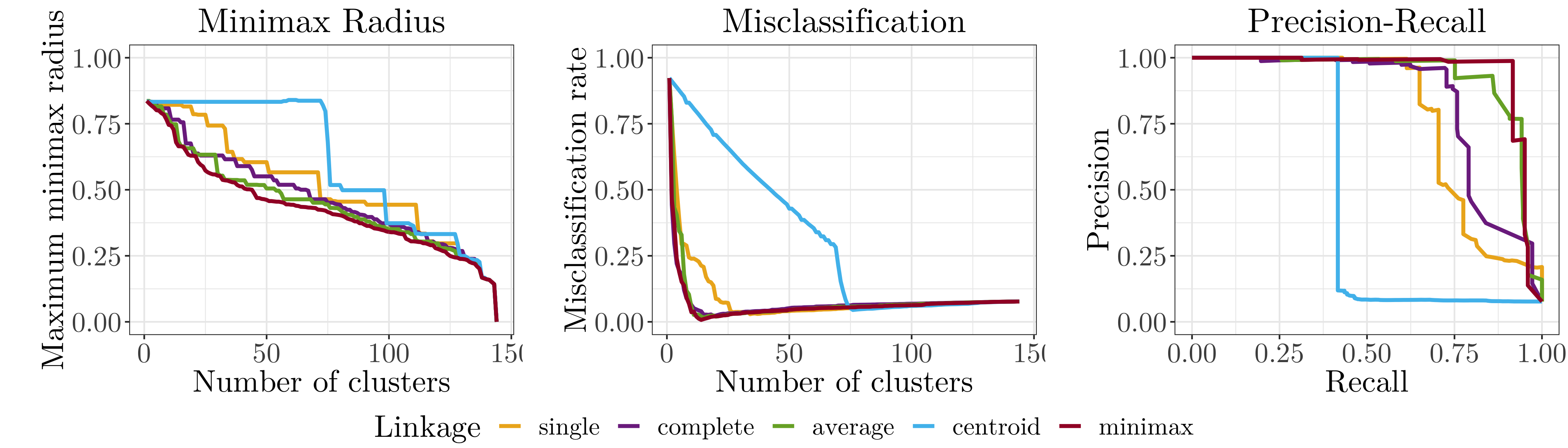}
\end{figure*}

\begin{figure*}[!ht]
	\caption{\label{fig:FBISW}
		Results for FBI S\&W study}
	\centering
	\includegraphics[width=\linewidth]{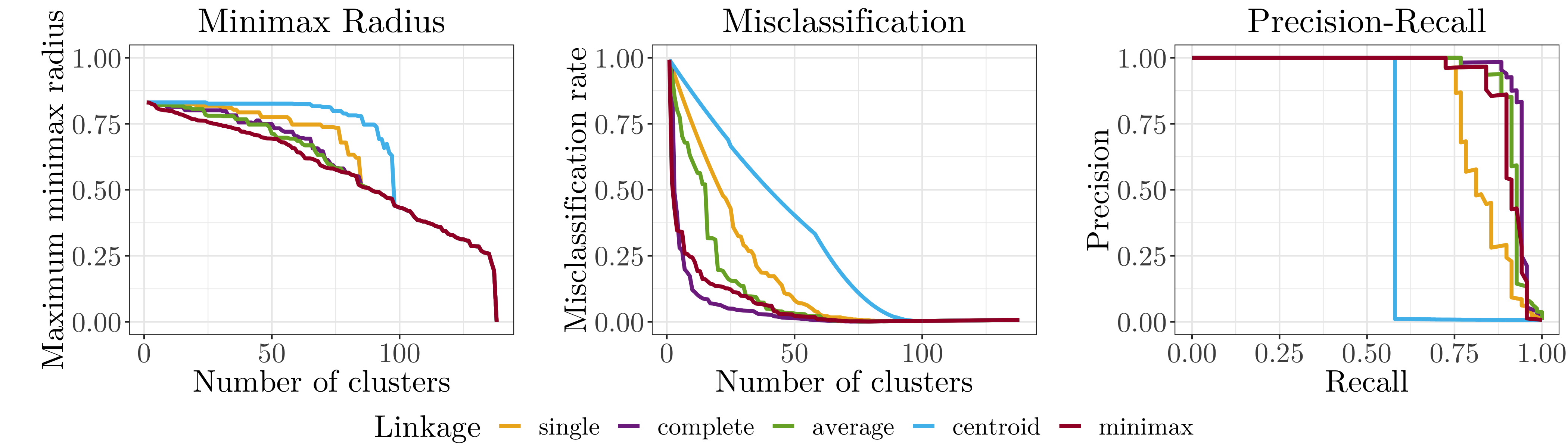}
\end{figure*} 

\end{document}